\documentclass{edm_article}

\usepackage{algorithm}
\usepackage{algorithmic}
\newlength\myindent
\newcommand\bindent{%
  \begingroup
  \setlength{\itemindent}{\myindent}
  \addtolength{\algorithmicindent}{\myindent}
}
\newcommand\eindent{\endgroup}
\begin{document}

\title{Auditing an Automatic Grading Model with deep Reinforcement Learning}

\numberofauthors{2}
 \author{
\alignauthor
Aubrey Condor\\
    \affaddr{University of California Berkeley}\\
    \email{aubrey\_condor@berkeley.edu}
\alignauthor
Zachary Pardos\\
    \affaddr{University of California Berkeley}\\
    \email{pardos@berkeley.edu}
}

\maketitle

\begin{abstract}
We explore the use of deep reinforcement learning to audit an automatic short answer grading (ASAG) model. Automatic grading may decrease the time burden of rating open-ended items for educators, but a lack of robust evaluation methods for these models can result in uncertainty of their quality. Current state-of-the-art ASAG models are configured to match human ratings from a training set, and researchers typically assess their quality with accuracy metrics that signify agreement between model and human scores. In this paper, we show that a high level of agreement to human ratings does not give sufficient evidence that an ASAG model is infallible. We train a reinforcement learning agent to revise student responses with the objective of achieving a high rating from an automatic grading model in the least number of revisions. By analyzing the agent's revised responses that achieve a high grade from the ASAG model but would not be considered a high scoring responses according to a scoring rubric, we discover ways in which the automated grader can be exploited, exposing shortcomings in the grading model.
\end{abstract}


\section{Introduction}

It has been shown that the use of open-ended (OE) items is beneficial for student learning due to the generation effect \cite{bertsch2007generation} or in combination with self-explanation \cite{chi1994eliciting}. However, assessing OE items is time consuming for teachers \cite{hancock1995implementing} and consequently, educators often default to using multiple choice (MC) questions. Automatic short answer grading (ASAG) models may help alleviate this time burden while encouraging educators to frequently incorporate OE items in their curriculum. To test the goodness of ASAG models, researchers typically follow the evaluation methods of machine learning (ML) researchers: using a hold-out or cross-validation approach where a portion of the data is used for training and another used for testing. Results are presented with metrics like accuracy or F1 score which signify a degree of agreement to the human ratings used to create the held-out test data. If an ASAG model shows concordance to human ratings, researchers generally say that the model performs well. However, we argue that even when training data are carefully constructed via meticulous grading from subject matter experts (SMEs), metrics which signify high agreement to an educator's ratings do not provide sufficient evidence that an ASAG model's scoring is robust. 

The purpose of this work is to audit an ASAG model by constructing unique responses that are granted high scores from a state-of-the-art ASAG model but would not receive a high score from an educator, according to a scoring rubric. We train a reinforcement learning (RL) agent to revise a student's response with the objective of obtaining a high score from an ASAG model in the least number of revisions. A revision consists of adding a phrase or deleting a portion of the students response. Because the RL agent has no knowledge of the scoring rubric and learns by trial-and-error over many thousands of revisions, we hypothesize that some of the high-scoring responses it learns to create would not qualify as a high scoring response according to a scoring rubric created by SMEs, providing evidence of flaws in the ASAG model. Although a human could find ways to exploit an ASAG model similarly through trial and error probing, the advantage of using RL is that the agent can create many thousands of responses quickly through its revisions, and does so specifically to achieve a high rating. We use a BERT-base multi-class classification model as our ASAG model which is a specific type of LLM \cite{devlin2018bert} that has been used extensively in recent ASAG research \cite{bonthu2021automated} \cite{camus2020investigating}.  The novelty of this research is not in training and evaluating the ASAG model, but in finding ways to probe a typically used ASAG model using RL.

We analyze results from three experimental settings, distinguished by which phrases the agent can add to a response. In all three experiments, the agent is able to delete a section of the student's response in the same way. To define the phrases that the agent is able to add to a response, we reference the scoring rubric that was created by subject matter experts for the original human ratings. In the first experiment, the agent is able to insert any of 10 key phrases which each represent an idea determined to be important for achieving a high score as specified in the rubric. We set the agent up for success by giving it choices that are likely to increase a response's rating. In the second experiment, we set the agent up for failure as the agent can add any of 10 unhelpful phrases that should not increase the probability of achieving a high rating, but that include relevant terminology. In the third experiment, the agent can add any of 20 phrases including the 10 key phrases from the first experiment and the 10 unhelpful phrases from the second experiment. 

\section{Related Work}
\subsection{ASAG with Deep Learning}
Recent ASAG research exploits advances in deep learning methods. Haller et al. (2023) provide a comprehensive analysis of recently published methods which deploy deep learning approaches for ASAG. They emphasize that the progression from engineered features of text to representation learning methods, including the use of word embeddings, sequential models and attention-based methods, has contributed to recent improvements in ASAG models \cite{haller2022survey}. Further, much of the newest ASAG work makes use of large language models (LLMs). Camus and Filighera compared the performance of LLMs for ASAG in terms of the size of the transformer and the ability to generalize to other languages \cite{camus2020investigating}, and Condor et al. explored whether a BERT ASAG model may be improved by further pre-training on graphical representations of scoring rubrics \cite{condor2022representing} . Agarwal et al. (2022) introduced the Multi-Relational Graph Transformer (MitiGaTe) to prepare token representations that consider the structural context of a sentence \cite{agarwal2022multi}, and Zhu et al. (2022) proposed a novel BERT-based, NN framework for ASAG by adjusting BERT embeddings with a semantic refinement layer consisting of a bidirectional-Long Short-Term Memory (LSTM) network and a Capsule network with position information \cite{zhu2022automatic}. Additionally, researchers are investigating the robustness of such grading models: Filighera et al. (2020) found that ASAG systems can be fooled with universal adversarial trigger employment \cite{filighera_steuer_rensing_2020}, and Condor et al. (2021) attempted to make LLM ASAG models more generalizable to out-of-training-sample questions by incorporating question-related text \cite{condor2021automatic}.

\subsection{RL for Education}
The use of RL for applications in education is not yet broadly explored. A survey article by Singla et al. summarized that the main contributions in RL for education have been personalizing curriculum, providing hints, adaptive experimentation, modeling students, and content generation \cite{singla2021reinforcement}. Notable works include Fazazi et al. who presented a multi-agent approach to deep Q-larning for designing an adaptive e-Learning system which customizes to a particular learner’s characteristics  \cite{el2021design}. MacLellan and Gupta used Proximal Policy Optimization to learn expert models for educational tasks and compare their RL approach to existing computational approaches \cite{maclellan2021learning}. Doroudi et al. reviewed a variety of RL based attempts at instructional sequencing, and concluded that RL has been most successful in when it is constrained by theories from the psychological sciences \cite{doroudi2019s}. Finally, Cai et al. presented MathBot, a bandit algorithm chatbot that explains math concepts, provides practice questions, and offers personalized feedback to students \cite{cai2021bandit}, and Condor and Pardos \cite{condor2022deep} created an RL algorithm to provide formative feedback to students about their open-ended responses in a simulated environment. 

\section{Background}
\subsection{Deep Reinforcement Learning}
RL employs learning to control a dynamical system by providing feedback to an agent via a reward system, such that the agent learns to choose an optimal action in a state. The action consists of what the agent can do in the state, and the state represents the current condition of the agent. When the agent takes an action, it receives a reward as feedback from the environment. The agent’s goal is to learn a policy to maximize its total expected sum of rewards \cite{sutton2018reinforcement}. Traditionally, RL algorithms implement online learning by iteratively collecting experiences while actively interacting with an environment, but offline RL algorithms utilize previously collected data. Our policy is trained using offline learning because as the agent will make many mistakes while it learns, and “bad” revisions could hinder a student’s learning progression. Our RL agent is trained via Proximal Policy Optimization - a family of policy gradient methods for reinforcement learning \cite{sutton2018reinforcement}.

\subsection{The ASAG Model}
The ASAG model that we use to provide ratings for student responses as well as feedback to the agent through a reward function is a BERT-base multi-class classification model. BERT is a popular language model that has been used frequently in ASAG research. The BERT transformer language model, introduced in \cite{devlin2018bert}, can be fine-tuned for downstream tasks such as classification. We use a compressed version of the model called BERT-base due to computational resource limitations. The model is fine-tuned as a supervised classifier using 781 human ratings of the OE question(s) as ground-truth training ratings, just as in previous ASAG works such as \cite{condor2020exploring}. The model was trained using a train-validation-test split of 70\%, 15\%, 15\% respectively. The held-out test set achieved a Quadratic Weighted Kappa of 0.7919 and a ROC AUC value of 0.9185

\section{Methods}
\subsection{The Reinforcement Learning Formulation}
The agent's training objective is to iteratively revise a response to achieve a high grade from the ASAG model in the least number of revisions. For each episode, we will initialize by randomly sampling a previously collected student response. Based on the response, the agent will choose an action to take via the current policy, i.e., either adding a phrase or deleting a portion of the response. Next, we send the newly revised student response to the ASAG model and receive a probability distribution of rating categories. We calculate the reward as a function of the difference in rating class probabilities from the previous response to the newly revised response, and if the reward has surpassed a predefined threshold, the episode will terminate. Otherwise, the process is repeated with the revised student response until the reward threshold is achieved, or the agent reaches a maximum number of revisions. Further details follow. 

\begin{algorithm}
\begin{algorithmic}
\caption{The RL Formulation}
\STATE Repeat (for each episode):
    \bindent
    \STATE Randomly sample student response
    \STATE Get response rating from ASAG model
    \STATE Repeat (for each time step):
        \STATE\hspace{\algorithmicindent} Choose action and alter response 
        \STATE\hspace{\algorithmicindent} Get rating of altered response
        \STATE\hspace{\algorithmicindent} Calculate reward
    \STATE Until max steps or rating threshold reached
    \eindent
\end{algorithmic}
\end{algorithm}

\textbf{The State} consists of the response that the agent is currently revising. For a given student's response, if the agent has already made a revision, the state would consist of the revised response (i.e., the original response plus the added text or without the deleted text). \textbf{The Action Space} consists of the phrases that the agent may add and where it chooses to place the phrase within the response - either in the front (before the first word) or at the end (after the last word) of the current response. The agent may also choose to delete any 1/5th sized section of the existing response. Thus, the action space is size 25 for the first two experiments ((2 locations * 10 phrases) + 5 delete options), and size 45 for the third experiment ((2 locations * 20 phrases) + 5 delete options). We only allow the agent to add phrases to two locations and limit the flexibility of its' delete actions to keep the action space at a reasonable size. The maximum number of revisions the agent can make in one episode (to one response) is eight. \textbf{The Policy} is a BERT-base model because of the model's impressive ability to learn representations of text. The input to the BERT policy is the state, and output is a probability distribution over the action space. \textbf{The Reward} is a scaled difference in the expected value of the old rating to the expected value of the new rating, with a penalty for each revision the agent makes. We include the penalty because we want the agent to learn how to revise the student's response in the least number of revision steps. We scale the difference in expected values by multiplying by 3 because the overall difference tended to be small and we needed to send a stronger signal to the agent. The reward formula is as such:

$ 3 \cdot (\sum_{x=0}^{4} x_{new} \cdot p(x_{new}) - \sum_{x=0}^{4} x_{old} \cdot p(x_{old})) -1 $

\subsection{The Data Set}
We used a data set from a previous research project at UC Berkeley consisting of OE science items designed for middle school students \cite{riordan2020empirical}. Responses were scored with a Knowledge Integration (KI) rubric by subject matter experts. KI is a framework for strengthening science understanding that emphasizes incorporating new ideas and sorting out alternative perspectives with evidence \cite{linn2000designing}. For this project, we use one item from a unit about the physics of sound waves that engages students to refine their ideas about concepts like wavelengths, frequency and pitch. Students must distinguish how the pitch of sound made by tapping a full glass of water compares to the pitch made by tapping an empty glass \cite{riordan2020empirical}. The data include 1313 OE responses rated from 1 (lowest rating) to 5 (highest rating). A detailed rubric was created by researchers and used for scoring that includes a description of the rating level, examples of correct/incorrect mechanisms and conclusions, and examples of student responses that would fall into each category.

\subsection{The Experiments}
We investigate how the RL agent finds ways to exploit the ASAG model by defining three different experiments. In all three experiments, the agent can delete any 1/5th response length section as 5 of its actions. What differs between the experiments is the set of phrases the agent can choose to insert either before the first word of the response or after the last word of the response. In the first experiment, the agent may insert any of 10 key phrases into a response. These phrases were taken directly from the SME-created rubric as important mechanisms or conclusions that a student could include in their response to achieve a high rating. In the second experiment, we include 10 phrases that include words related to the subject matter, but that do not contain information that should lead to a higher score. We will refer to these as unhelpful phrases. In the third experiment, we allow the agent to insert any of 20 phrases including both the important phrases and unhelpful phrases. Below are the phrases for the first two experimental setups. \\
\\
\textbf{Experiment 1 - Helpful Phrases:}
\textit{'sound moves faster in air', 'sound moves slower in water', 'water is more dense', 'water has more mass', 'higher frequency in air', 'pitch is lower in water', 'pitch higher in empty glass', 'air is less dense', 'less vibration in water', 'the pitch is different'}
\\
\textbf{Experiment 2 - Unhelpful Phrases:}
\textit{'I am not sure', 'tapping the glass', 'sound bounces in the glass', 'sound sinks in water', 'sound will echo in glass', 'the pitch is the same', 'water blocks the sound', 'frequency is height of wave', 'amplitude is number of waves', 'sound is more dense'}

\section{Results}
For each experiment, we train the agent 10 times for 75,000 time steps, differing only in the random initialization of the policy network. Each training run yields a different number of episodes depending on how many revisions it takes the agent to arrive at the threshold reward, so we analyze the first 18,250 episodes of each run. Visually, we can compare the mean episode return over time, smoothed with a rolling average of n=200, of the 10 runs across each experiment in Figure 1. We include 95\% confidence bands for each smoothed mean as well. It is apparent that the training runs for experiment 1 (Helpful Phrases) and experiment 3 (All Phrases) consistently yield higher average episode returns than those for experiment 2 (Unhelpful Phrases). We are not surprised by this result as less revisions correspond to a higher return, and we hypothesized that the agent should be able to achieve the high threshold rating with a lesser number of revisions using helpful phrases than with the unhelpful phrases. We also see significant overlap in the confidence bands between experiment 1 (Helpful Phrases) and experiment 3 (All Phrases). In the sections that follow, we provide results of an exploratory data analysis of action sequences for the top 5\% of episode returns for each of the three experiments. \textbf{When showing examples of student responses and corresponding revisions, we show added phrases in brackets, and deleted pieces of a response with a strike-through.} 

\begin{figure}[t]
\centering
\includegraphics[width=9cm]{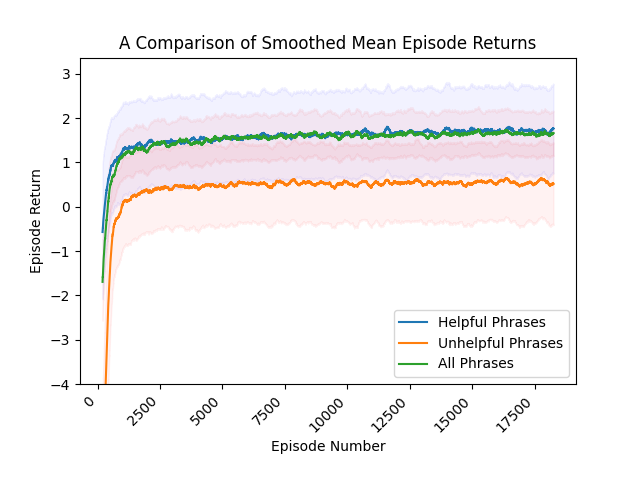} 
\caption{Mean returns of the first 18,250 episodes over 10 training runs for each experiment, smoothed with a rolling average of n=200.}
\label{fig1}
\end{figure}

\subsection{Experiment 1: Helpful Phrases}
We analyze 9,125 total high scoring revision sequences that correspond to the top 5\% of returns over all 10 training runs in experiment 1. The mean number of actions necessary to achieve a high ASAG rating is 3.4. Interestingly, the agent rarely - only 21 sequences out of 9,125 (0.23\%) - chooses to delete a section of a response in the top scoring sequences. The most compelling finding is that the agent frequently chooses to add the same phrase multiple times within a revision sequence - the number of sequences with at least one repeated phrase addition is 2,353 or 26\% of all 9,125 sequences we analysed. Further, in 161 revision sequences, the agent chooses to add the same phrase at least three times in a row (1.76\% of the time). \textbf{For example}: "[water has more mass] in a pool its lower and slower a glass wouldnt be different [water has more mass] [water has more mass] [water has more mass]" (after 4 revisions)

\subsection{Experiment 2: Unhelpful Phrases}
We analyze 9,146 total high scoring revision sequences that correspond to the top 5\% of returns over all 10 training runs in experiment 2. The mean number of actions necessary to achieve a high ASAG rating is 3.12. Although the top 5\% of scores for experiment two is lower than that of experiment 1 (as is evident in Figure 1), we emphasize that the revised responses in the top 5\% of experiment two are still able to achieve the high threshold rating from the ASAG model, using the supposedly unhelpful phrases. The most interesting finding is that 84.3\% of actions chosen across all revision sequences correspond to adding the phrase, \textit{"sound is more dense".} Further, the phrase, \textit{"frequency is height of wave"} is chosen 14.8\% of the time. Even more pronounced than that of experiment 1 is the repeating of the same action choice multiple times within a revision sequence - this happens in 7,916 total revision sequences (86.6\%) and the majority of these cases consist of adding the \textit{"sound is more dense"} phrase repeatedly. \textbf{For example}: "the waves travel farther making a higher pitch [sound is more dense] [sound is more dense]" (after 2 revisions)

\subsection{Experiment 3: All Phrases}
We analyze 9,125 total high scoring revision sequences that correspond to the top 5\% of returns over all 10 training runs in experiment 3. The mean number of actions necessary to achieve a high ASAG rating is 3.5. We find that 90\% of all actions taken in the top 5\% consist of adding 'Helpful Phrases' from experiment 1, and only 10\% consist of adding 'Unhelpful Phrases' from experiment 2. Of the actions that the agent chooses from the 'Unhelpful Phrases', results are consistent to that of experiment 2 in that by far, the most frequently chosen phrase to add was \textit{"sound is more dense"}. This phrase was chosen about 95\% of the time an unhelpful phrase was chosen. Further, consistent with experiments 1 and 2, we see the agent repeats the same phrase addition at least once within a sequence in 2172 (24\%) of the 9,125 revisions analysed. \textbf{For example}: "[pitch is lower in water] i dont know [water has more mass] [water has more mass] [sound is more dense]" (After 4 revisions) 

\section{Discussion}
In all three experiments, it appears that the agent is able to increase ASAG response scores by adding the same phrase multiple times. This should certainly not be the case as repeating a phrase does not indicate an increase in understanding. Further, the idea that the agent is able to improve a response's automatic grade with any of the unhelpful phrases is evidence that there are issues to be resolved with the ASAG model. We noticed that agent disproportionately favored incorrect phrases like \textit{"sound is more dense"} which is a typical misunderstanding of students, as specified in the scoring rubric. The ASAG model seems to be "tricked" by this phrase, as it includes relevant terminology like "more dense". Finally, evident from Experiment 3, although the agent highly favors the helpful phrases, it does choose the unhelpful phrases about 10\% of the time.

Limitations of our work include that our analysis is dependent on the action spaces we chose for the agent in each experiment. There may be a better or different assortment of unhelpful phrases that would have been useful to include in the action space for experiment two. Further, our analysis was specific to a ASAG BERT model and although this is a particularly popular option for recent ASAG research, many researchers are using different variants of LLMs like RoBERTa or Llama. Additionally, we analyze results from one specific question related to science understanding. Finally, as our analysis was exploratory, it was only reasonable to find general patterns in the agent's revisions as opposed to examining each high scoring revision sequence.

We have found a few interesting flaws in a frequently used, state-of-the-art, ASAG model. For this work to be practically useful, we hope that these findings may be helpful for improving similar LLM ASAG models. As we continue this work, we plan to use these undesirable examples to retrain the ASAG model. For instances where the agent added incorrect phrases, or inserted the same phrase multiple times, we can rate those responses as lower than their original, un-revised scores, and add the altered responses and ratings to the original data set to retrain the ASAG model. We would hypothesis that the retrained ASAG model may not reward responses with repeated phrases or non-useful phrases. We have shown that there are still issues to be sorted out with such models and importantly, the creative use of deep RL may help education researchers search for such flaws in similar scenarios with large search spaces.

\section{Acknowledgments}
This work was supported in part by funding from the UC Berkeley Peder Sather Center for Advanced Study.

\bibliographystyle{abbrv}
\bibliography{RL_EDM}  
%

\balancecolumns
\end{document}